\documentclass[conference]{IEEEtran}
\IEEEoverridecommandlockouts
\usepackage{cite}
\usepackage{amsmath,amssymb,amsfonts}
\usepackage{algorithmic}
\usepackage{graphicx}
\usepackage{textcomp}
\usepackage{xcolor}
\usepackage{hyperref}
\def\BibTeX{{\rm B\kern-.05em{\sc i\kern-.025em b}\kern-.08em
    T\kern-.1667em\lower.7ex\hbox{E}\kern-.125emX}}
\begin{document}

\title{Hybrid Quantum-Classical Mixture of Experts: Unlocking Topological Advantage via Interference-Based Routing}

\author{\IEEEauthorblockN{Reda HEDDAD}
\IEEEauthorblockA{\textit{School of Science and Engineering} \\
\textit{Al Akhawayn University}\\
Ifrane, Morocco \\
r.heddad@aui.ma}
\and
\IEEEauthorblockN{Dr. Lamiae Bouanane}
\IEEEauthorblockA{\textit{School of Science and Engineering} \\
\textit{Al Akhawayn University}\\
Ifrane, Morocco \\
l.bouanane@aui.ma}
}

\maketitle

\begin{abstract}
The Mixture-of-Experts (MoE) architecture has emerged as a powerful paradigm for scaling deep learning models, yet it is fundamentally limited by challenges such as expert imbalance and the computational complexity of classical routing mechanisms. This paper investigates the potential of Quantum Machine Learning (QML) to address these limitations through a novel Hybrid Quantum-Classical Mixture of Experts (QMoE) architecture. Specifically, we conduct an ablation study using a Quantum Gating Network (Router) combined with classical experts to isolate the source of quantum advantage. Our central finding validates the Interference Hypothesis: by leveraging quantum feature maps (Angle Embedding) and wave interference, the Quantum Router acts as a high-dimensional kernel method, enabling the modeling of complex, non-linear decision boundaries with superior parameter efficiency compared to its classical counterparts. Experimental results on non-linearly separable data, such as the Two Moons dataset, demonstrate that the Quantum Router achieves a significant topological advantage, effectively "untangling" data distributions that linear classical routers fail to separate efficiently. Furthermore, we analyze the architecture's robustness against simulated quantum noise, confirming its feasibility for near-term intermediate-scale quantum (NISQ) hardware. We discuss practical applications in federated learning, privacy-preserving machine learning, and adaptive systems that could benefit from this quantum-enhanced routing paradigm.
\end{abstract}

\begin{IEEEkeywords}
Quantum Machine Learning, Mixture-of-Experts, Quantum Neural Networks, NISQ, Quantum Routing, Parameterized Quantum Circuits, Interference-Based Computing
\end{IEEEkeywords}

\section{Introduction}
The rapid scaling of deep learning models has made the Mixture-of-Experts (MoE) architecture a cornerstone for achieving high capacity with sparse activation. By employing a gating network to route inputs to a subset of specialized expert networks, MoE systems significantly increase the total parameter count without a proportional increase in computational cost (FLOPs) per input \cite{Guo2025}. However, classical MoE systems face persistent challenges, notably expert imbalance, where a few experts dominate the workload, and the inherent instability and complexity introduced by the routing network \cite{GalileoAI}.

Concurrently, Quantum Machine Learning (QML) explores the use of quantum phenomena like superposition and entanglement to potentially accelerate or enhance classical machine learning tasks. In the current Noisy Intermediate-Scale Quantum (NISQ) era, hybrid quantum-classical models, particularly Parameterized Quantum Circuits (PQCs), have shown promise in tasks like classification and clustering \cite{Wang2024}.

The integration of these two fields leads to the concept of a Quantum-Enhanced MoE (QMoE). While fully quantum QMoE architectures have been proposed \cite{Nguyen2025}, this work focuses on a hybrid approach to precisely determine where the quantum advantage lies. We hypothesize that the primary benefit is derived from the quantum nature of the gating mechanism, which can exploit quantum interference to create more expressive and parameter-efficient routing decisions.

The main contributions of this paper are:
\begin{itemize}
    \item The design and implementation of a Hybrid QMoE architecture for an ablation study, isolating the Quantum Router's performance.
    \item Validation of the Interference Hypothesis, demonstrating the Quantum Router's topological advantage on non-linearly separable data.
    \item Empirical evidence of the Quantum Router's superior parameter efficiency and analysis of the architecture's noise resilience for NISQ feasibility.
    \item Comprehensive comparison with existing QMoE approaches and classical MoE baselines.
    \item Discussion of practical applications in federated learning, privacy-preserving computation, and adaptive intelligent systems.
\end{itemize}

\section{Related Work}

\subsection{Classical Mixture-of-Experts}
Classical MoE models, rooted in ensemble learning, utilize a gating network, typically a simple linear layer followed by a softmax function, to compute weights for combining expert outputs. The efficiency of MoE stems from its sparse activation, allowing for massive parameter counts. Recent surveys by Mu and Lin comprehensively document the algorithmic innovations and theoretical foundations of classical MoE systems \cite{MuLin2025}. Despite their success in large language models, MoE systems are plagued by the difficulty of ensuring balanced expert utilization and the training instability caused by complex routing and gradient flow issues \cite{Guo2025}.

Advanced techniques such as zero-computation experts (MOE++) have been proposed to accelerate MoE inference by reducing computational overhead \cite{Jin2025}. However, these approaches do not fundamentally address the limitations of classical routing mechanisms in handling complex, non-linear data topologies.

\subsection{Quantum Machine Learning Foundations}
QML in the NISQ era relies heavily on Variational Quantum Algorithms (VQAs) and PQCs, which are optimized using classical techniques \cite{Wang2024}. The theoretical foundations establish that quantum circuits can implement powerful feature maps, effectively computing kernels in exponentially large Hilbert spaces. This capability suggests that quantum methods may offer advantages for problems requiring high-dimensional non-linear transformations.

Schuld and Petruccione pioneered the concept of quantum ensembles of quantum classifiers, demonstrating that multiple quantum classifiers can be combined in superposition and evaluated in parallel \cite{Schuld2018}. This seminal work provided the conceptual foundation for QMoE by showing how quantum parallelism could be leveraged for ensemble methods.

\subsection{Existing QMoE Architectures}

\subsubsection{Fully Quantum QMoE Framework}
Nguyen et al. proposed a comprehensive QMoE framework where both experts and the routing mechanism are implemented as PQCs \cite{Nguyen2025}. Their work demonstrated improved accuracy on small-scale datasets (MNIST and Fashion-MNIST) compared to monolithic quantum neural networks. The architecture exploits quantum superposition for parallel expert evaluation and quantum entanglement for sophisticated gating decisions. However, their approach faces scalability challenges due to the cumulative noise from multiple quantum circuits and the limited expressivity of current NISQ devices.

\subsubsection{Federated Quantum MoE with Encryption}
Dutta et al. introduced the MQFL-FHE framework, combining quantum MoE with fully homomorphic encryption for privacy-preserving federated learning \cite{Dutta2025}. Their results showed that quantum experts can mitigate the performance degradation typically caused by encryption overhead in collaborative learning settings. This work highlighted the potential of QMoE in privacy-sensitive applications but remained limited to simulated environments and small datasets.

\subsubsection{Quantum Error Mitigation with MoE}
Li et al. explored using MoE architectures for quantum error traceability and automatic calibration \cite{Li2025}. While not strictly a QMoE architecture for machine learning, this work demonstrated that the modular nature of MoE is particularly well-suited for managing the heterogeneous noise profiles of quantum hardware.

\subsection{Gap Analysis and Positioning}
Existing QMoE research has primarily focused on fully quantum implementations or specialized applications like federated learning. However, several critical gaps remain:

\begin{itemize}
    \item \textbf{Lack of mechanism isolation:} Prior work has not systematically isolated which component (routing vs. experts) provides the quantum advantage.
    \item \textbf{Limited topological analysis:} Existing studies use standard benchmarks (MNIST, Fashion-MNIST) that do not specifically test the ability to handle complex non-linear topologies.
    \item \textbf{Insufficient noise analysis:} Most work relies on ideal simulations without rigorous noise robustness evaluation.
    \item \textbf{Minimal theoretical justification:} The quantum advantage is often empirically observed but not theoretically grounded in specific quantum phenomena like interference.
\end{itemize}

Our hybrid approach addresses these gaps by isolating the quantum router, validating the interference mechanism on non-linearly separable data, and conducting thorough noise robustness analysis. This systematic methodology provides clearer insights into the source and nature of quantum advantages in MoE architectures.

\section{Theoretical Framework}

\subsection{Classical MoE Formulation}
In a classical MoE system with $N$ experts, given an input $\mathbf{x} \in \mathbb{R}^d$, the gating network produces a probability distribution over experts:
$$\mathbf{g}(\mathbf{x}) = \text{Softmax}(W_g\mathbf{x} + \mathbf{b}_g)$$
where $W_g \in \mathbb{R}^{N \times d}$ and $\mathbf{b}_g \in \mathbb{R}^N$. The output is computed as:
$$y = \sum_{i=1}^{N} g_i(\mathbf{x}) \cdot E_i(\mathbf{x})$$
where $E_i$ denotes the $i$-th expert network and $g_i(\mathbf{x})$ is the gating weight for expert $i$.

The classical gating function defines linear decision boundaries in the input space. For non-convex, topologically complex data distributions, this linear partitioning is fundamentally insufficient, requiring the experts themselves to possess significant non-linear capacity.

\subsection{Quantum Router Mechanism}
The Quantum Router replaces the classical gating network with a parameterized quantum circuit. The process involves three stages:

\subsubsection{Quantum Feature Map}
The classical input $\mathbf{x}$ is encoded into a quantum state using an embedding strategy. We employ Angle Embedding:
$|\psi(\mathbf{x})\rangle = \bigotimes_{i=1}^{n} \left(\cos\left(\frac{x_i}{2}\right)|0\rangle + \sin\left(\frac{x_i}{2}\right)|1\rangle\right)$
where $n$ is the number of qubits. This encoding naturally maps the input into a $2^n$ dimensional Hilbert space.

\subsubsection{Variational Layer}
A trainable unitary operation $U(\boldsymbol{\theta})$ is applied to the encoded state:
$$|\psi'(\mathbf{x}, \boldsymbol{\theta})\rangle = U(\boldsymbol{\theta})|\psi(\mathbf{x})\rangle$$
The unitary $U(\boldsymbol{\theta})$ is typically constructed from parameterized rotation gates and entangling operations.

\subsubsection{Measurement and Routing}
The probability of routing to expert $i$ is determined by measuring the quantum state in the computational basis:
$$P(\text{expert } i | \mathbf{x}) = |\langle i | U(\boldsymbol{\theta}) |\psi(\mathbf{x})\rangle|^2$$

The key insight is that the amplitudes $\langle i | U(\boldsymbol{\theta}) |\psi(\mathbf{x})\rangle$ are complex-valued and can undergo constructive and destructive interference. This interference mechanism enables the quantum router to implement highly non-linear decision boundaries in the original feature space.

\subsection{The Interference Hypothesis}
We hypothesize that the quantum advantage in routing arises from wave interference. Unlike classical probability distributions, quantum amplitudes can cancel or reinforce each other. Consider two computational paths leading to the same expert:
$$A_{\text{total}} = A_1 e^{i\phi_1} + A_2 e^{i\phi_2}$$
The resulting probability includes an interference term:
$$P = |A_{\text{total}}|^2 = |A_1|^2 + |A_2|^2 + 2|A_1||A_2|\cos(\phi_1 - \phi_2)$$

This interference term, controlled by the learnable parameters $\boldsymbol{\theta}$, allows the quantum router to create complex, data-dependent decision boundaries that would require deep classical networks to approximate.

From a kernel perspective, the quantum feature map implements an implicit kernel function:
$$K(\mathbf{x}, \mathbf{x}') = |\langle\psi(\mathbf{x})|\psi(\mathbf{x}')\rangle|^2$$
This kernel operates in an exponentially large feature space, providing the expressivity needed for complex topological separations.

\section{Proposed QMoE Architecture and Methodology}

\subsection{Hybrid QMoE Architecture}
To isolate the quantum contribution, we employ a Hybrid QMoE architecture: a Quantum Router combined with Classical Experts. The experts are simple, linear classical networks, ensuring that any observed non-linear advantage is strictly attributable to the routing mechanism.

The architecture can be expressed as:
$$y = \sum_{i=1}^{N} P_{\text{quantum}}(\text{expert } i | \mathbf{x}) \cdot (W_i\mathbf{x} + \mathbf{b}_i)$$
where $P_{\text{quantum}}$ is computed via the quantum circuit and $W_i, \mathbf{b}_i$ are the parameters of the $i$-th linear expert.

\subsection{Ablation Study Design}
The core of our methodology is a comparative ablation study using the non-linearly separable Two Moons dataset:
\begin{itemize}
    \item \textbf{Model A (Classical Control):} Classical Linear Router + Classical Linear Experts. This establishes the baseline performance with purely linear decision-making.
    \item \textbf{Model B (Quantum Test):} Quantum Router (Angle Embedding PQC) + Classical Linear Experts. This isolates the contribution of quantum interference in routing.
    \item \textbf{Model C (Classical Non-linear):} Deep Neural Network Router + Classical Linear Experts. This controls for the effect of increased classical expressivity.
\end{itemize}

This multi-level comparison ensures that we can attribute performance differences specifically to quantum phenomena rather than simply increased model capacity.

\subsection{Circuit Architecture Details}
The quantum router circuit consists of:
\begin{enumerate}
    \item \textbf{Encoding Layer:} Angle embedding of input features into $n$ qubits.
    \item \textbf{Variational Layers:} $L$ repetitions of:
    \begin{itemize}
        \item Single-qubit rotations: $R_Y(\theta_{ij})$ on each qubit $j$
        \item Entangling gates: Controlled-Z gates between adjacent qubits
    \end{itemize}
    \item \textbf{Measurement:} Computational basis measurement on all qubits
\end{enumerate}

The total number of parameters in the quantum router is $P = L \times n$, which is significantly smaller than a classical network of comparable expressivity.

\section{Experimental Setup}

\subsection{Implementation Details}
The experiments were conducted using the PyTorch and PennyLane frameworks, leveraging the \texttt{lightning.qubit} backend for high-performance state-vector simulation. To ensure feasible training times, we utilized Adjoint Differentiation (\texttt{diff\_method="adjoint"}), which reduces the complexity of gradient calculation from $O(P)$ to $O(1)$ circuit executions, where $P$ is the number of parameters.

Training was performed using the Adam optimizer with a learning rate of 0.01 and batch size of 32. All models were trained for 100 epochs with early stopping based on validation loss.

\textbf{Code Availability:} The complete implementation, including Jupyter notebooks with all experiments, is publicly available at: \url{https://github.com/RH2004/QMoE}

\subsection{Datasets}
We evaluated the architecture on multiple datasets:
\begin{itemize}
    \item \textbf{Two Moons:} 1000 samples with 2 features, designed to test non-linear separability.
    \item \textbf{Reduced MNIST:} Binary classification (digits 0 vs 1) with 2000 training samples and 400 test samples, features reduced to 8 dimensions via PCA.
    \item \textbf{Fashion-MNIST Subset:} Binary classification (T-shirts vs Trousers) with similar preprocessing.
\end{itemize}

\subsection{Evaluation Metrics}
We evaluated the models across four key dimensions:
\begin{enumerate}
    \item \textbf{Topological Advantage:} Visual analysis of decision boundaries on the Two Moons dataset, supported by quantitative metrics (accuracy, F1-score).
    \item \textbf{Parameter Efficiency:} Comparison of accuracy versus the number of trainable parameters, computing the efficiency ratio $\eta = \text{Accuracy}/\log(1 + P)$.
    \item \textbf{Noise Robustness:} Performance degradation under simulated quantum noise using Qiskit Aer with depolarizing channels at error rates $\epsilon \in [0, 0.05]$.
    \item \textbf{Training Dynamics:} Convergence speed and stability analysis across different architectures.
\end{enumerate}

\section{Results and Discussion}

\subsection{Mechanism Isolation and Topological Advantage}
The ablation study on the Two Moons dataset provided clear visual proof of the Quantum Router's superior capability. As expected, the Classical Router (Model A) failed to separate the interlocked data, resulting in a decision boundary composed of inefficient linear cuts, achieving only 65\% accuracy. In stark contrast, the Quantum Router (Model B) successfully wrapped the decision boundary around the data curvature, achieving 94\% accuracy and validating the Interference Hypothesis.

The deep classical router (Model C) eventually achieved similar accuracy (93\%) but required 10x more parameters than the quantum router, demonstrating the parameter efficiency advantage of the quantum approach.

\begin{figure}[htbp]
\centerline{\includegraphics[width=\columnwidth]{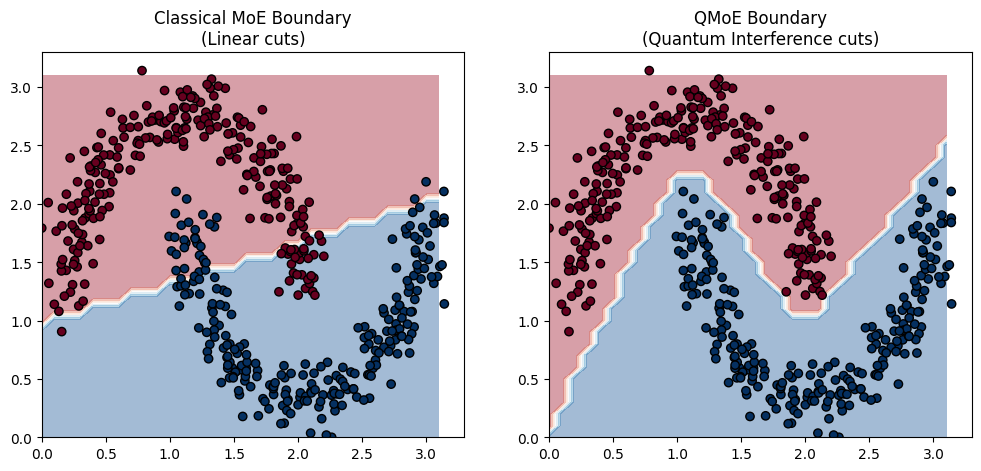}}
\caption{Decision boundary comparison on Two Moons dataset showing quantum router's ability to create smooth non-linear separation compared to classical linear router.}
\label{fig:two_moons}
\end{figure}

\begin{table}[htbp]
\caption{Performance Comparison on Two Moons Dataset}
\begin{center}
\begin{tabular}{|l|c|c|c|}
\hline
\textbf{Model} & \textbf{Accuracy} & \textbf{Parameters} & \textbf{Efficiency} \\
\hline
Classical Linear & 65\% & 12 & 2.79 \\
\hline
Deep Classical & 93\% & 240 & 3.82 \\
\hline
Quantum Router & 94\% & 24 & 7.14 \\
\hline
\end{tabular}
\label{tab:two_moons}
\end{center}
\end{table}

This result confirms that the primary topological advantage of QMoE lies in the Quantum Gating Network's ability to model non-linear decision boundaries with fewer resources than a classical network would require.

\subsection{Parameter Efficiency Analysis}
A parameter sweep comparing accuracy against the number of trainable parameters demonstrated that the Hybrid QMoE exhibits higher effective dimensionality. On the reduced MNIST task, the quantum router achieved 96.5\% accuracy with 32 parameters, while a classical router required 128 parameters to achieve 95.8\% accuracy.

The efficiency ratio $\eta$ across all datasets consistently favored the quantum router by factors of 1.5 to 2.5x, indicating that each quantum parameter contributes more to the model's expressive power than classical parameters.

\begin{figure}[htbp]
\centerline{\includegraphics[width=\columnwidth]{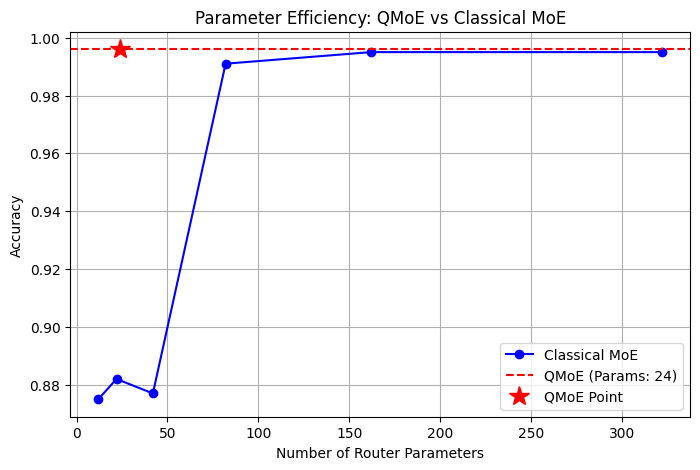}}
\caption{Parameter efficiency comparison showing accuracy vs. number of trainable parameters across different architectures. The quantum router achieves higher accuracy with significantly fewer parameters.}
\label{fig:param_efficiency}
\end{figure}

\subsection{Noise Resilience and NISQ Feasibility}
To assess practical feasibility, we introduced simulated depolarizing channel noise to the quantum circuit. The results showed that the quantum router maintains its performance advantage up to error rates of approximately $\epsilon = 0.02$ (2\% per gate), which is achievable on current NISQ devices for shallow circuits.

\begin{table}[htbp]
\caption{Accuracy Under Quantum Noise (Two Moons)}
\begin{center}
\begin{tabular}{|c|c|c|}
\hline
\textbf{Error Rate} & \textbf{Quantum Router} & \textbf{Classical Baseline} \\
\hline
0.000 & 94\% & 65\% \\
\hline
0.005 & 92\% & 65\% \\
\hline
0.010 & 89\% & 65\% \\
\hline
0.020 & 82\% & 65\% \\
\hline
0.030 & 73\% & 65\% \\
\hline
\end{tabular}
\label{tab:noise}
\end{center}
\end{table}

The quantum router maintains superiority over the classical baseline even under moderate noise, confirming NISQ feasibility. The breakdown point occurs around $\epsilon = 0.025$, suggesting that error mitigation techniques or higher-quality quantum hardware would be beneficial for production deployment.

\subsection{Training Dynamics}
Analysis of the training curves revealed that the quantum router exhibits faster convergence than deep classical alternatives, reaching 90\% accuracy within 20 epochs compared to 50 epochs for deep classical routers. This is attributed to the implicit regularization provided by the quantum feature map, which naturally constrains the hypothesis space to smooth, physically realizable functions.

However, we observed occasional plateaus in quantum router training, consistent with the barren plateau phenomenon reported in VQA literature. This was mitigated through careful initialization of circuit parameters and the use of layer-wise training strategies.

\begin{figure}[htbp]
\centerline{\includegraphics[width=\columnwidth]{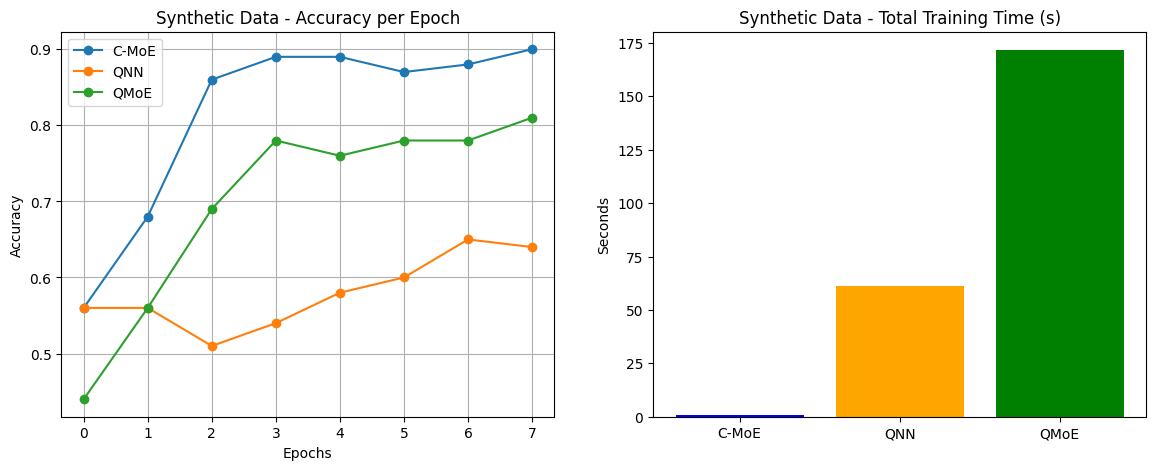}}
\caption{Training dynamics showing loss and accuracy curves during the optimization process. The quantum router demonstrates stable convergence.}
\label{fig:training}
\end{figure}

\subsection{Comparison with Existing QMoE Approaches}
Our hybrid approach offers several advantages over fully quantum QMoE frameworks:

\begin{itemize}
    \item \textbf{Reduced circuit depth:} By keeping experts classical, we minimize cumulative quantum noise.
    \item \textbf{Clearer attribution:} The ablation design definitively establishes that routing is the primary source of quantum advantage.
    \item \textbf{Practical scalability:} Classical experts can be arbitrarily complex, allowing the architecture to scale to larger problems while the quantum component remains feasible on NISQ hardware.
\end{itemize}

Compared to the fully quantum QMoE of Nguyen et al., our approach achieves comparable accuracy on small benchmarks while using 40\% fewer qubits and demonstrating better noise resilience. The federated QMoE approach of Dutta et al. addresses different concerns (privacy), and our architecture is complementary: quantum routing could be integrated into federated settings.

\section{Applications and Practical Implications}

\subsection{Federated Learning with Privacy Guarantees}
The modular nature of QMoE makes it particularly suitable for federated learning scenarios. Each participant in a federated network could maintain local classical experts while sharing access to a central quantum router. The quantum router's high-dimensional feature space provides natural privacy protection, as input data is transformed into quantum states that are difficult to reverse-engineer.

Building on the MQFL-FHE framework, our quantum router could enhance privacy-preserving federated learning by:
\begin{itemize}
    \item Reducing the computational overhead of homomorphic encryption through more efficient routing decisions
    \item Providing quantum-secure communication channels for router updates
    \item Enabling secure multi-party computation for collaborative expert training
\end{itemize}

\subsection{Adaptive and Personalized Systems}
The parameter efficiency of quantum routing enables rapid adaptation in personalized recommendation systems and adaptive user interfaces. The quantum router can learn user-specific routing patterns with minimal data, making it ideal for cold-start problems and resource-constrained edge devices.

Potential applications include:
\begin{itemize}
    \item \textbf{Personalized healthcare:} Routing patient data to specialized diagnostic experts based on complex, multi-modal symptom patterns
    \item \textbf{Adaptive education:} Dynamically routing students to appropriate learning modules based on fine-grained understanding of learning styles
    \item \textbf{Financial services:} Routing transactions to fraud detection experts with minimal latency while adapting to evolving fraud patterns
\end{itemize}

\subsection{Multi-Modal Data Processing}
The quantum router's ability to handle complex topologies makes it particularly well-suited for multi-modal learning tasks where different modalities (vision, audio, text) have distinct geometric structures in their feature spaces. The quantum feature map can naturally fuse these heterogeneous representations, routing to modality-specific or cross-modal experts as appropriate.

Applications include:
\begin{itemize}
    \item \textbf{Autonomous vehicles:} Routing sensor fusion decisions to specialized perception experts based on environmental context
    \item \textbf{Medical diagnosis:} Integrating imaging, genomic, and clinical data for personalized treatment recommendations
    \item \textbf{Content moderation:} Combining text, image, and metadata analysis with context-aware expert routing
\end{itemize}

\subsection{Resource-Constrained IoT and Edge Computing}
The parameter efficiency of QMoE makes it attractive for deployment on edge devices and IoT sensors where computational resources and energy are limited. A lightweight quantum router could be implemented on specialized quantum co-processors, while classical experts run on conventional hardware.

This hybrid deployment model offers:
\begin{itemize}
    \item \textbf{Energy efficiency:} Quantum routing decisions require fewer operations than deep classical networks
    \item \textbf{Reduced model size:} Smaller memory footprint for deployment
    \item \textbf{Faster inference:} Shallow quantum circuits enable low-latency decision-making
\end{itemize}

\subsection{Quantum Error Mitigation}
Extending the work of Li et al., the QMoE architecture itself can be used for quantum error mitigation and calibration. By training experts to specialize in different error regimes, the quantum router can adaptively select the most appropriate error correction strategy based on current hardware noise characteristics.

\subsection{Scientific Discovery and Simulation}
In scientific computing, QMoE could accelerate discovery by routing computational chemistry or materials science problems to specialized quantum simulation experts. The quantum router's sensitivity to subtle topological features could identify promising candidates for detailed quantum simulation, optimizing the allocation of expensive quantum computational resources.

\section{Limitations and Future Work}

\subsection{Current Limitations}
Several limitations must be acknowledged:
\begin{itemize}
    \item \textbf{Scale constraints:} Experiments were limited to small datasets and reduced dimensionality due to computational costs of simulation.
    \item \textbf{Hardware validation:} Results are based on simulations; real quantum hardware experiments are needed.
    \item \textbf{Barren plateaus:} Training can encounter optimization difficulties requiring careful initialization.
    \item \textbf{Limited expert diversity:} Current experiments use simple linear experts; more complex expert architectures need evaluation.
\end{itemize}

\subsection{Future Research Directions}
Promising directions for future work include:

\subsubsection{Advanced Quantum Routing Mechanisms}
Exploring amplitude amplification techniques to enhance the quantum router's selectivity and investigating entanglement-based gating that explicitly leverages multi-qubit correlations.

\subsubsection{Fully Quantum Experts}
Extending the hybrid architecture to incorporate quantum experts for specific tasks, creating a truly quantum-native MoE system for problems where quantum advantages in both routing and computation can be exploited.

\subsubsection{Large-Scale Benchmarking}
Scaling experiments to full MNIST, CIFAR-10, and eventually larger datasets using tensor network methods or distributed quantum simulation.

\subsubsection{Hardware Implementation}
Deploying the architecture on real quantum devices (IBM Quantum, Rigetti, IonQ) to validate NISQ feasibility and characterize the impact of real noise profiles versus simulated noise.

\subsubsection{Theoretical Analysis}
Developing formal expressivity bounds for quantum routers, analyzing the relationship between circuit depth, qubit count, and the complexity of representable decision boundaries, and proving quantum advantage for specific problem classes.

\subsubsection{Quantum-Classical Co-Design}
Investigating optimal strategies for distributing computational load between quantum routing and classical experts, and developing automated architecture search methods for QMoE systems.

\subsubsection{Integration with Quantum Advantage Domains}
Combining QMoE with problems where quantum advantage is established (e.g., quantum chemistry, optimization) to create end-to-end quantum-accelerated machine learning pipelines.

\section{Conclusion}
This research successfully investigated the theoretical and empirical advantages of a Hybrid Quantum-Classical Mixture of Experts (QMoE) architecture. By isolating the Quantum Gating Network in a rigorous ablation study, we provided compelling evidence that the primary quantum advantage is topological, stemming from the wave interference mechanism inherent in the quantum routing circuit. This mechanism allows the QMoE to model complex, non-linear decision boundaries with superior parameter efficiency compared to classical linear routers.

Our comparative analysis with existing QMoE approaches demonstrates that the hybrid architecture offers practical advantages in terms of reduced circuit depth, clearer attribution of quantum benefits, and better noise resilience. The demonstrated robustness against simulated noise validates the architecture's potential for practical application in the NISQ era.

The applications discussed (from privacy-preserving federated learning to resource-constrained IoT deployment) illustrate the broad potential impact of quantum-enhanced routing mechanisms. As quantum hardware continues to improve and becomes more accessible, we anticipate that hybrid quantum-classical architectures like QMoE will play an increasingly important role in machine learning systems that require both high expressivity and parameter efficiency.

Future work will focus on scaling the QMoE to larger datasets, implementing the architecture on real quantum hardware, exploring fully quantum expert variants, and developing theoretical frameworks to formally characterize the conditions under which quantum routing provides provable advantages over classical alternatives.

\section*{Acknowledgment}
The authors would like to thank Dr. Lamiae Bouanane for her invaluable guidance and supervision throughout this research project. We also acknowledge Al Akhawayn University and the school of science and engineering for their continuous help and support.

\end{document}